\documentclass[a4paper, 10pt, conference]{ieeeconf}      
\usepackage{FG2020}
\usepackage{epsfig}
\usepackage{graphicx}
\usepackage{amsmath}
\usepackage{amssymb}
\usepackage[utf8]{inputenc} 
\usepackage[T1]{fontenc}    
\usepackage{url}            
\usepackage{booktabs}       
\usepackage{amsfonts}       
\usepackage{nicefrac}       
\usepackage{microtype}      
\usepackage{bm}
\usepackage{bbm}
\usepackage{times}
\usepackage{epsfig}
\usepackage{algorithm}
\usepackage{algpseudocode}
\usepackage{xcolor,colortbl}
\usepackage{graphicx}
\usepackage{color}
\usepackage{stmaryrd}

\FGfinalcopy 

\IEEEoverridecommandlockouts                              
\def\FGPaperID{****} 

\title{\LARGE \bf
Deep Entwined Learning Head Pose and Face Alignment Inside an Attentional Cascade with Doubly-Conditional fusion
}

\author{\parbox{16cm}{\centering
    {\large Arnaud Dapogny$^{1,2}$, Kevin Bailly$^{1,3}$ and Matthieu Cord$^{2}$}\\
    {\normalsize
    $^1$ Datakalab, 114 boulevard Malesherbes, 75017 Paris\\
    $^2$ LIP6, Sorbonne Universit\'e, CNRS, 4 place Jussieu, 75005 Paris\\
    $^3$ ISIR, Sorbonne Universit\'e, CNRS, 4 place Jussieu, 75005 Paris}}
    \thanks{This work was supported by the French National Agency (ANR) in the frame of its Technological Research JCJC program (FacIL, project ANR-17-CE33-0002).}
}

\begin{document}

\ifFGfinal
\thispagestyle{empty}
\pagestyle{empty}
\else
\author{Anonymous FG2020 submission\\ Paper ID 71\\}
\pagestyle{plain}
\fi
\maketitle

\begin{abstract}
Head pose estimation and face alignment constitute a backbone preprocessing for many applications relying on face analysis. While both are closely related tasks, they are generally addressed separately, e.g. by deducing the head pose from the landmark locations. In this paper, we propose to entwine face alignment and head pose tasks inside an attentional cascade. This cascade uses a geometry transfer network for integrating heterogeneous annotations to enhance landmark localization accuracy. Furthermore, we propose a doubly-conditional fusion scheme to select relevant feature maps, and regions thereof, based on a current head pose and landmark localization estimate. We empirically show the benefit of entwining head pose and landmark localization objectives inside our architecture, and that the proposed AC-DC model enhances the state-of-the-art accuracy on multiple databases for both face alignment and head pose estimation tasks.
\end{abstract}

\section{Introduction}
\label{sec:intro}

Face alignment refers to the process of localizing a number of landmarks on the face, such as lips or eye corners, pupils or nose tips. It serves as a backbone preprocessing for many applications in computer vision, such as expression recognition \cite{dapogny2015pairwise}, face synthesis \cite{thies2016face2face}, or facial performance reenactment \cite{thies2016face2face}. Depending on the application, the precision of the annotation markup varies a lot, \textit{i.e.} the number of annotated landmarks can be very different across datasets. How to integrate these heterogeneous annotations in order to robustly localize variable numbers of landmarks belonging to different markups remains an open challenge.

A closely related task is head pose estimation, which is usually formulated as a regression task on the three Euler angles (yaw, pitch and roll). Given a precise landmark localization, head pose can be estimated quite straightforwardly. However, there is no guarantee that such two-step multi-task formulation is optimal at all: for example, given a rough head pose estimate, one can theoretically select more relevant face features in order to specialize landmark localization to a certain pose range, as the face appearance varies wildly between e.g. yaw angles close to $90$ degrees and smaller angles, particularly when considering cheek regions.

\begin{figure}[ht]
	\centering
	\includegraphics[width=0.9\linewidth]{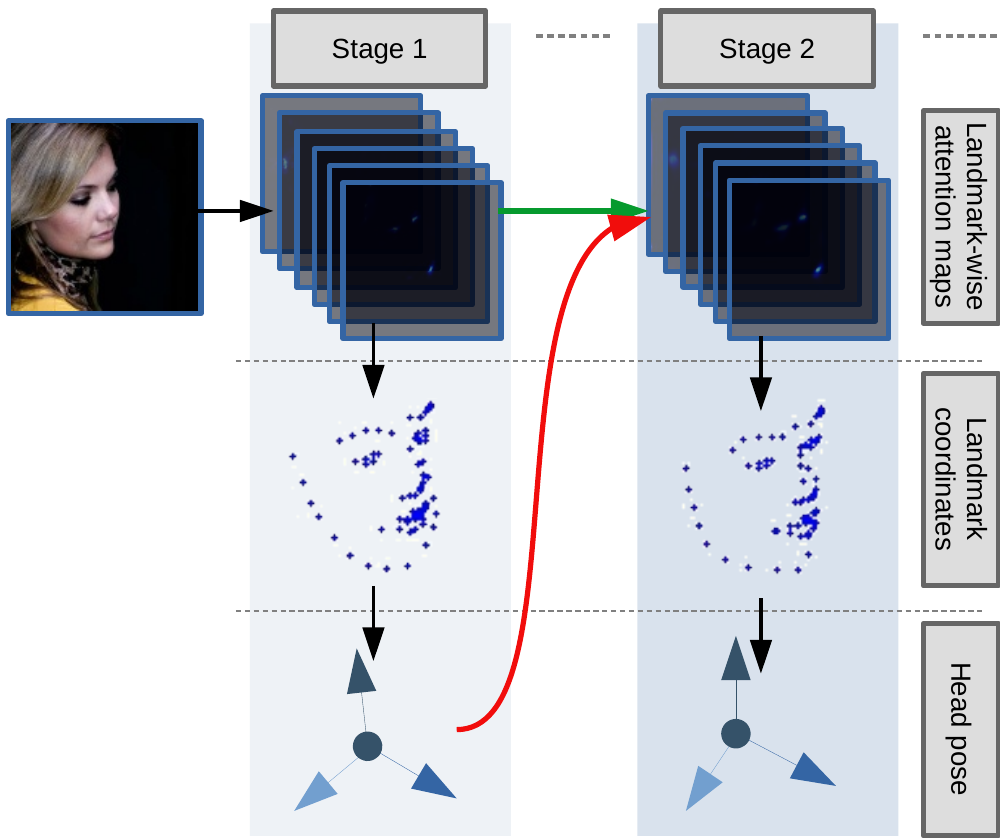}
	\caption{Illustration of how face alignment and head pose estimation are entwined in AC-DC. Facial landmarks are jointly estimated in the first cascade stage, then used to estimate head pose. Both landmark-wise attention maps (green arrow) and head pose estimate (red arrow) are then used \textit{via} doubly-conditional fusion to refine each other through the subsequent stages.}
	\label{illu}
\end{figure}

\begin{figure*}[h!]
	\centering
	\includegraphics[width=\linewidth]{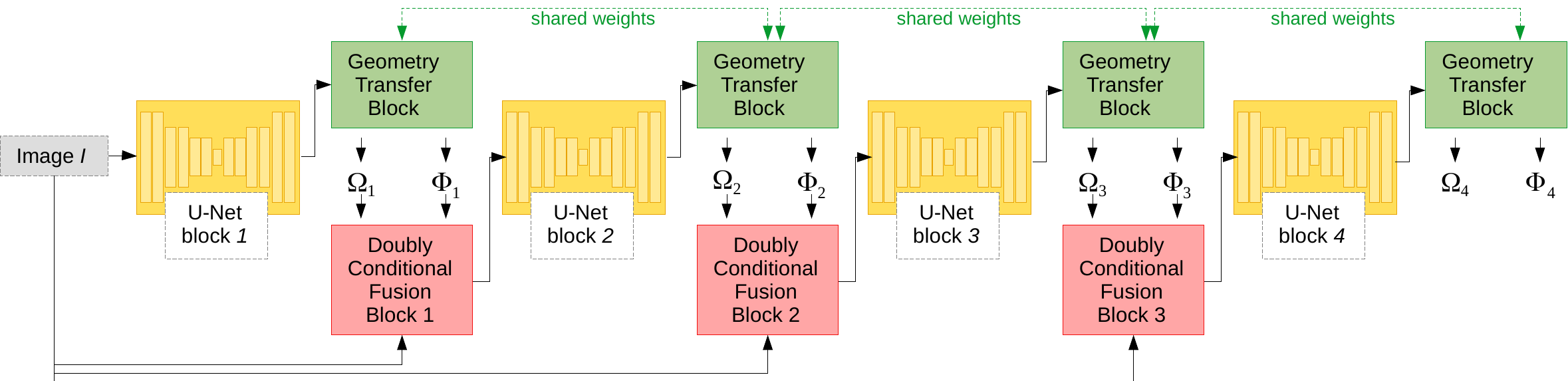}
	\caption{AC-DC is composed of four stacked blocks composed of a 13-layers fully-convolutional U-Net as well as a doubly conditional feature fusion block.}
	\label{main}
\end{figure*}

In this paper, we propose to entwine head pose estimation and face alignment tasks, so that each task benefit from each other. This process is illustrated on figure \ref{illu}. To do this, we introduce an attentional cascade with doubly conditional fusion (AC-DC) model. AC-DC is an hybrid architecture between cascaded regression methods and deep learning-based end-to-end learnable approaches. It is composed of several stages that each contains a backbone fully-convolutional U-net block. For each stage, a geometry transfer network (GTN) converts the current landmark estimates to fit multiple, heterogeneous annotation markups, and provides a current head pose estimate. Then, a doubly-conditional fusion block uses both head pose estimates and spatial attention maps to select relevant channels, and regions thereof, to provide richer embeddings to the next cascade stages. The whole architecture is trained in an end-to-end fashion. To wrap it up, the contributions of this paper are three-folds:
\begin{itemize}
\item We propose an attentional cascade (AC-DC) that iteratively refines head pose and landmark estimates. It uses a  doubly conditional pose and spatial masking to refine the predictions of the next stages.
\item We propose a dual-stream geometry transfer network (GTN) to integrate heterogeneous landmark and head pose prediction objectives. 
\item We experimentally show that AC-DC extend state-of-the-art results by a significant margin for both 2D and 3D landmark alignment, as well as head pose estimation.
\end{itemize}

\section{Related work}

\textbf{Face alignment} can be divided into two coarse categories, the first of which is 2D alignment. This consists in predicting 2D landmark localization, usually for face images with low to medium yaw angles. Popular methods for 2D face alignment either belong to cascaded regression or deep learning-based, end-to-end approaches. Popular exemples of cascaded regression include SDM \cite{Xiong2013}, LBF \cite{Ren2014} and DAN \cite{kowalski2017deep}. A natural pitfall of such approaches is that the regressors are not learned jointly in a end-to-end fashion, thus there is no guarantee that the whole cascade might be optimal. Tackling this issue, MDM \cite{Trigeorgis2016} improves the feature extraction process by sharing CNN layers among cascade stages, which are formulated as a recurrent neural network. This results in a more optimized landmark trajectory throughout the cascade.

Exemples of deep methods include TCDCN \cite{zhang2016learning}, which involves pretraining on a wide facial attributes database \cite{liu2015deep}. More recently, SAN \cite{dong2018style} uses generative adversarial networks to convert images from different styles to an aggregated style before performing landmark localization. Authors of \cite{wu2018look} propose to use edge map estimation as an intermediate representation to drive the landmark prediction task. Authors in \cite{feng2018wing} use a surrogate loss to enhance training of deep networks. AAN \cite{yue2018attentional} proposes to use intermediate feature maps as attentional masks to select relevant regions.

3D face alignment methods are usually formulated as dense landmark localization objectives, which are degraded to a sparse set of landmarks for evaluation. For instance, PRN \cite{feng2018joint} learns a direct mapping between an input image and a UV map that contains 3D coordinates for each pixel. Such methods do not explicitly use head pose information. By contrast, 3DDFA \cite{zhu2019face} fits a 3D morphable model using a deep neural network. In such a case, head pose lies among the parameters of the morphable model and is explicitly estimated. However, such parametric model uses a restricted number of dimensions and is usually quite rigid, e.g. w.r.t. expressions.

\textbf{Head pose estimation:} like 3DDFA \cite{zhu2019face}, most methods in the literature either makes the assumption that head pose can be estimated prior to face alignment and can be used as a low-dimensional variable for conditioning the landmark localization task. This is due to the fact that head pose can be predicted accurately using a single deep network and without relying on landmark localization, as demonstrated in \cite{ruiz2018fine}. For instance, 3DDE \cite{valle2019face} first predicts a rough, rigid landmark localization guess using a deep neural network to estimate head pose, then refine its predictions with a coarse-to-fine ensemble of regression trees. PCD-CNN \cite{kumar2018disentangling} integrates head pose information inside a dentritic CNN architecture. Furthermore, many approaches such as \cite{werner2017landmark} treat head pose as a byproduct of the landmark localization, failing to enrich the latter task by the knowledge of pose estimation.

\textbf{Dealing with heterogeneous annotations:} besides head pose and landmark-wise annotations, face geometry is annotated in an heterogeneous fashion, with various numbers of landmarks. This is problematic, since fine-grained face annotation comes at a significant cost, hence available quality data is rather scarce. As an example, 300W database \cite{Sagonas2015} contains $\approx 3k$ images labelled in terms of 68 landmarks, whereas WFLW \cite{wu2018look} contains $7.5k$ images with 98 landmarks and CelebA \cite{liu2015deep} contains $\approx 200k$ images annotated with only 5 landmarks. Thus, one can wonder if we can use all those images within the same framework to learn more robust landmark predictions. This problem is usually tackled as a knowledge transfer between heterogeneous datasets \cite{zhang2015leveraging}. In \cite{wu2017leveraging} the sauthors use a multi-task formulation, with a separate regression head for every annotation markup. However, this essentially ignores the intrinsic relationship between the landmark alignment tasks. Finally, in prior work \cite{dapogny2019decafa} we proposed to chain landmark prediction tasks within a fully-convolutional attentional cascade. This approach, however, does not integrate 3D landmarks along with 2D annotations, nor do they entangle head pose with landmark localization.

\section{Attentional Cascade with Doubly-Conditional fusion}

An overview of AC-DC is illustrated on Figure \ref{main}. 4 fully-convolutional U-Net blocks are stacked on top of each other. A geometry transfer network (Section \ref{gtn}) converts these feature maps to address heterogeneous landmark alignment tasks, and provides a current estimate of head pose $\omega_i$ and landmark coordinates $s_i$. After each of these blocks, we apply our doubly conditional fusion block (Section \ref{dcf}) to produce the input for the subsequent U-net block. The whole architecture is trained in an end-to-end manner with intermediate supervisions (Section \ref{learning}) and hyperparameter settings specified in Section \ref{implemdetail}.

\subsection{Dual-stream geometry transfer network}\label{gtn}

\begin{figure}[ht]
	\centering
	\includegraphics[width=\linewidth]{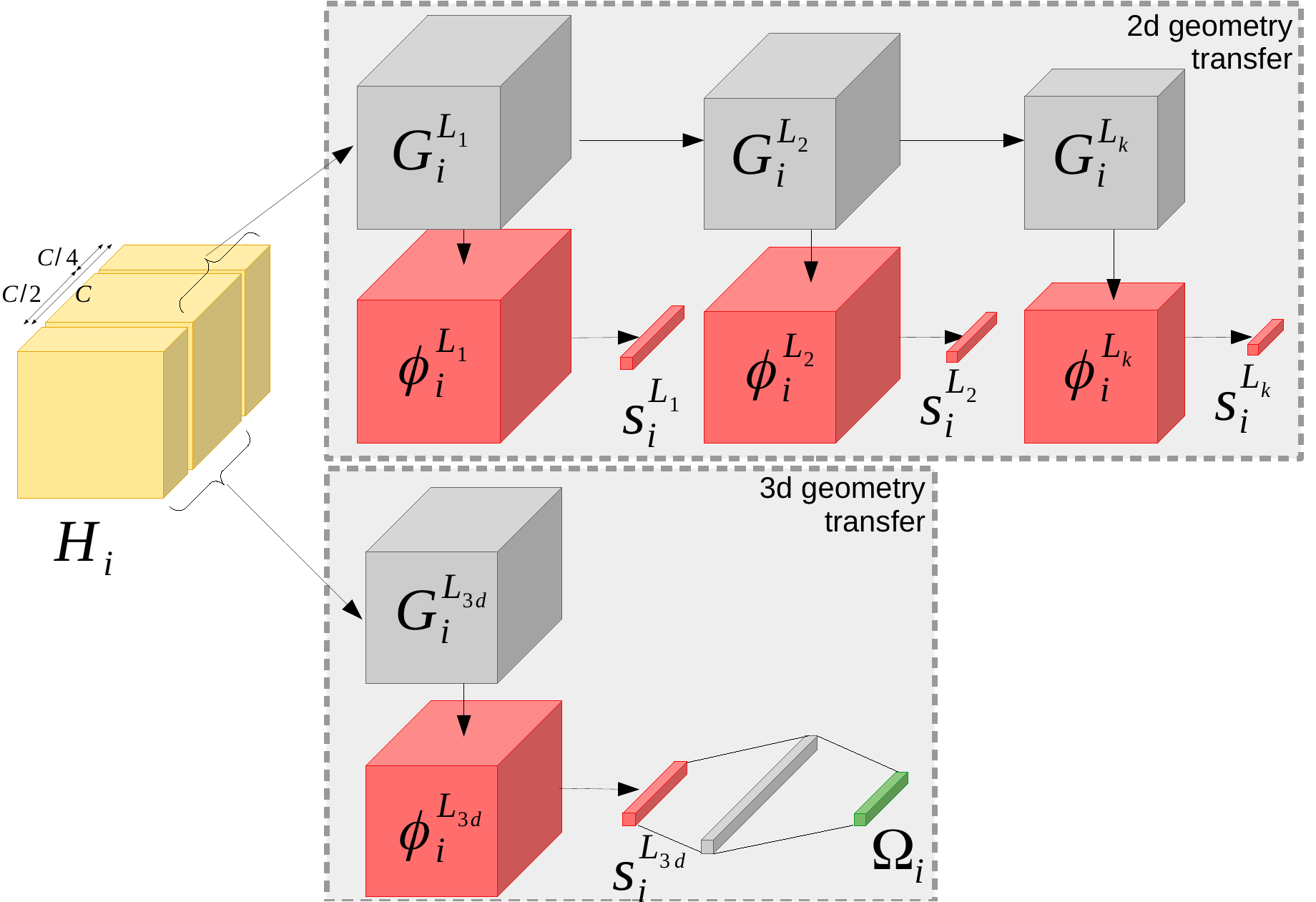}
	\caption{Illustration of the GTN. From the embeddings $H_i$, a number of channels are fed into a 2d geometry sub-network that convert these channels to landmark-wise pre-attention maps (gray boxes), attention maps (red) and landmark estimates by applying spatial softmax. A separated 3d geometry sub-network is also added.}
	\label{gtnfig}
\end{figure}

Let's denote $H_i$ the embeddings of the U-net block $i$, with $C=128$ channels. As we aim at aligning landmarks belonging to heterogeneous markups, we distinguish two cases: (a) markups that are semanticaly close to each other, or that can straightforwardly deduce from each other (such as \cite{Sagonas2015} and \cite{wu2018look}), and (b) semantically different markups (e.g. 2d \cite{Sagonas2015} and 3d landmarks). In order to integrate this constraint, we propose the dual-stream geometry transfer network (GTN) illustrated on Figure \ref{gtnfig}.
The GTN is composed of separate 2d and 3d sub-networks with distinct transfer layers. We use depthwise separable convolutions with overlap from $H_i$ so that the 2d and 3d alignment tasks benefit from each other while allowing the predicted landmarks to differ, most notably on landmarks belonging to the jawline. The 2d GTN contains several chained transfer layers for matching the number of landmarks of different markups with $L_1,...,L_k$ landmarks, respectively, ordered by ascending order $L_k \leq... \leq L_1$:

\begin{equation}
G^{L_k} = T^{L_k} \circ ... \circ T^{L_1} (H_i)
\end{equation}

With $T^{L_k}$ a $1 \times 1$ conv layer with $L_k$ output channels. By chaining these landmark pre-attention maps, each prediction task can benefit from the others, the finer tasks ($L_1$ landmarks) benefiting from the coarser ones ($L_k$ landmarks), as the gradients can flow from the former layers to the latter ones at train time. As for the 3d GTN, we only use one transfer layer as we only benchmark one 3d database in our experiments. From these 2d and 3d pre-attention maps $G^{L}$, we can derive landmark-wise attention maps by applying spatial softmax, that generates $L$ attention maps, one for each landmark:

\begin{equation}
\phi_i(x,y,l)=\frac{\exp ({G}_i^{L}(x,y,l))}{\sum \limits_{x=1}^X \sum \limits_{y=1}^Y \exp ({G}_i^{L} (x,y,l))}
\end{equation}

An estimation $s^L_i(l)$ of the $x-y$ coordinates for landmark $l$ of a $L$-landmarks markup can be obtained by computing the first order moments of $\phi_i(.,.,l)$:

\begin{equation}
s_i(l)=(\mathbb{E}_{x,y}[x\phi_i(x,y,l)],\mathbb{E}_{x,y}[y\phi_i(x,y,l)])
\end{equation}

This provides a differentiable estimate of the landmark coordinates. In particular, a head pose estimate $\Omega_i$ can be provided from $s^{L_{3d}}_i$ by applying a single dense layer. It should be noted that we use a single GTN by sharing the weights of all transfer layers, as well as the head pose estimation layer, between the different stages. The rationale behind doing this is that the geometric transformation that maps the different markups shall be intrinsic to the relationships between the tasks, hence not depending on a current estimation. 

\subsection{Doubly-Conditional Fusion Block}\label{dcf}

\begin{figure}[t]
	\centering
	\includegraphics[width=\linewidth]{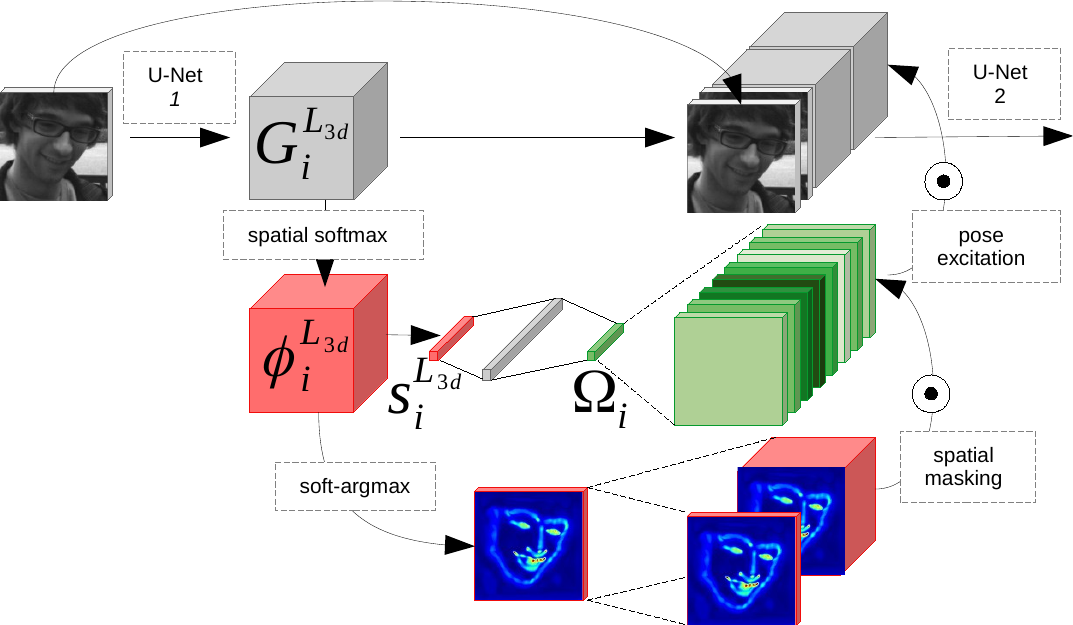}
	\caption{Illustration of our doubly-conditional fusion scheme. Head pose estimate, as well as a spatial mask obtained by aggregating landmark-wise attention maps, are used to select relevant channels, and regions thereof, to iteratively refine the network predictions.}
	\label{main2}
\end{figure}

Similarly to what is done in cascaded regression, we can then use the attention maps $\phi_i$ to select relevant regions for refining the landmark estimates. This is illustrated on Figure \ref{main2}. In order to limit the number of feature maps, we merge all the landmark-wise attention maps into a single spatial mask:

\begin{equation}
\Phi_i(x,y)= \max_l \Big[ \frac {\phi_i(x,y,l) - \min_{x,y} \phi_i(x,y,l)}{\max_{x,y} \phi_i(x,y,l) - \min_{x,y} \phi_i(x,y,l)} \Big]
\end{equation}

The spatially-conditional fusion output $F_{AC}$ of an attentional cascade (AC) can thus be written:

\begin{equation}
F_i^{AC}=I \oplus (I \odot \Phi_i) \oplus H_i \oplus (H_i \odot \Phi_i)
\end{equation}

Where $\oplus$ denotes channel concatenation and $\odot$ the Hadamard product. We can use the head pose estimate to emphasize the most relevant channels among the $130$ channels of $F_i^{AC}$, that shall be used by the next block. To do so, we map $\Omega_i$ to a 130-dimensional output $f(\Omega_i)$ by applying a fully-connected layer with sigmoid activation. Thus the output of the doubly-conditional fusion block $F_i^{AC-DC}$ is:

\begin{equation}
F_i^{AC-DC}=f(\Omega_i) \odot F_i^{AC}
\end{equation}

Where $\odot$ indicates the Hadamard product replicated along the spatial dimensions of the embeddings. This is similar to Squeeze-and-Excitation \cite{hu2018squeeze}. However, we constrain the squeeze layer to describe head pose information, which is notoriously relevant for landmark alignment \cite{dantone2012real}. Also note that while the parameters GTN (including the head pose estimation layer) are shared across all cascade stages, the parameters of the pseudo-excitation layer $f:\Omega_i \rightarrow f(\Omega_i)$ are not shared to allow to select different channels depending on the cascade stage $i$. In what follows, we call this model the attentional cascade with doubly conditional (AC-DC) network. Note that AC-DC is fairly deep ($\approx 60$ convolutional layers), but contains less that $10M$ parameters, thus is relatively light as compared to state-of-the-art approaches.

\subsection{Training AC-DC model}\label{learning}

AC-DC is trained in an end-to-end manner by minimizing a $\mathcal{L}_1$ loss between the predicted landmark locations and pose, and their ground truth counterpart: $\theta^*=argmin_\theta \{ \mathcal{L}_s(\theta) + \mathcal{L}_\Omega(\theta) \}$ where $\theta$ denotes the parameters of the 4 U-net backbone and GTN, as well as head pose excitation layers. In order to facilitate training, similarly to what is traditionally done in cascaded approaches, we drive the learning of each stage with intermediate supervision using both landmark and head pose ground truth values with:

\begin{equation}\label{lossfun}
\mathcal{L}_s(\theta)= \sum \limits_{i=1}^4 \lambda_i \sum \limits_{k=1}^K \frac{1}{L_k} |s_i^{L_k}-s^{L_k*}|
\end{equation}

the landmark localization objective function, and

\begin{equation}\label{lossfun}
\mathcal{L}_\Omega(\theta)= \sum \limits_{i=1}^4 \lambda_i |\Omega_i-\Omega^*|
\end{equation}

the head pose estimation term, and $\lambda_i$ denoting the intermediate supervision weights. In what follows, we use ascending weights $\lambda_1=0.125,\lambda_2=0.25,\lambda_3=0.5,\lambda_4=1$ to enable proper cascaded alignment as suggested in prior work \cite{dapogny2019decafa}, with the first stages outputting coarse predictions that are refined throughout the attentional cascade.

\subsection{Implementation details}\label{implemdetail}

In what follows, we use 4-stages AC-DC models that takes as input $128 \times 128$ grayscale images. Each U-Net block is composed of a $1 \times 1$ convolution layer, an encoder and a decoder part. The input of the $1 \times 1$ conv is the original $128 \times 128 \times 1$ grayscale image for block 1, and $128 \times 128 \times 130$ tensor for blocks 2,3,4. The encoder part of each block performs subsequents applications of $3 \times 3$ conv, batch norm, ReLU, followed $3 \times 3$ conv with stride 2, batch norm, ReLU. The number of channels is $64 \rightarrow 64 \rightarrow 128 \rightarrow 128 \rightarrow 256 \rightarrow 256$. The decoder part mirrors the encoding part. In order to generate smooth feature maps we do not use transposed convolution but instead use bilinear image upsampling followed with $3 \times 3$ convolutional layers. Furthermore, skip connections are used between feature maps of the same size to preserve the full spatial resolution of the input image. The whole architecture is trained using ADAM optimizer with a $5e^{-4}$ learning rate with $\beta_1=0.9$ and learning rate annealing with power $0.9$. We apply $400000$ updates with batch size $8$ for each database, with alternating updates between the databases.

\section{Experiments}

In this Section, we validate our models on several databases and metrics specified in Section \ref{expsetup}. We first highlight the benefits of our doubly-conditional fusion scheme in Section \ref{ablstudy}, then proceed to compare AC-DC with recent approaches for 2d alignment in Section \ref{2dal}, 3d alignment in Section \ref{3dal}, and head pose in Section \ref{hppred}. Finally, in Section \ref{visus} we show qualitative results obtained with AC-DC.

\subsection{Experimental setup}\label{expsetup}

The \textbf{300W} database, introduced in \cite{Sagonas2015}, contains moderate variations in pose and expressions. It also embraces a few occluded images. It consists in four databases: \textbf{LFPW} (811 images for train / 224 images for test), \textbf{HELEN} (2000 images for train / 330 images for test), \textbf{AFW} (337 images for train) and \textbf{IBUG} (135 images for test), for a total of 3148 images annotated with 68 landmarks for training the models. As state-of-the-art approaches already outputs very high accuracy on this dataset, authors of \cite{zhu2016face} introduced the \textbf{300W-LP} database, which is a large-pose dataset synthesized from 300W. It contains 100842 train images and 21608 images following the same partitions as in 300W, but with yaw angles covering the $[-90,90]$ degrees range. Authors of \cite{zhu2016face} also proposed AFLW2000-3D database, which contains example synthesized from the 2000 first images of AFLW database using the same protocol as in 300W-LP.

The \textbf{CelebA} database \cite{liu2015deep} is a large-scale face attribute database which contains $202599$ $218 \times 178$ celebrity images coming from $10177$ identities, each annotated with $40$ binary attributes (such as \textit{gender}, \textit{eyeglasses}, \textit{smile}), and the localization of $5$ landmarks (nose, left and right pupils, mouth corners). In our experiments, we use the train partition that contains $162770$ images from $8k$ identities to train our models. The test partition contains $19962$ instances from $1k$ identities that are different from the training set identities.

The \textbf{Wider Facial Landmarks in the Wild or WFLW} database \cite{wu2018look} contains 10000 faces (7500 for training and 2500 for testing) with 98 annotated landmarks. This database also features rich attribute annotations in terms of occlusion, head pose, make-up, illumination, blur and expressions.

In our experiments, we train on the train partitions of 300W, 300W-LP, WFLW, and CelebA and evaluate our models on the test partitions of these datasets as well as AFLW2000-3D. We report three evaluation metrics, the normalized mean error (NME), the failure rate or FR@0.1 and the AUC@0.1. For 2d alignment, the NME denotes the average landmark-wise distance normalized by the inter-ocular distance (distance between the outer eye corners). For 3d face alignment, as it is traditionnally done in the literature, we normalize the distances using the square root of the bounding box $height \times width$, as proposed in \cite{zhu2019face}. The FR@0.1 corresponds to the proportion of examples for which the NME is larger than 0.1, and AUC@0.1 is the integral or the cumulative error distribution (CED) curve for examples for which the NME is below 0.1. For head pose estimation, we report the mean absolute difference (MAE) for each Euler angle, as well as the average error over these angles.

\subsection{Ablation study}\label{ablstudy}

In this Section, we discuss the interest of using a doubly-conditional fusion. For validation of hyperparameters such as the number of cascade stages, task ordering and intermediate supervision weights, the reader shall refer to \cite{dapogny2019decafa}. In Table \ref{ablation4} we show the interest of our doubly conditional fusion scheme on WFLW database. First, we measure the performance of a simply-conditionnal attentional cascade (AC). In this case we use only spatial masking and no excitation layer to produce the input on each stage. By contrast, AC+Pose is obtained by adding a head pose loss (Equation \eqref{lossfun}), but without using head pose as an excitation variable to select relevant channels to refine the embeddings for the next U-net block. As such, using AC+Pose already provides an improvement over AC on every subset. Furthermore, by selecting relevant channels using head pose information (AC-DC), we improve the landmark localization accuracy of the model on nearly every subset of the database, most notably on the head pose and make-up subsets. This validates the fact that entwining head pose within the landmark alignment task by applying doubly-conditional fusion improves the performance of the model for landmark localization, particularly in case of difficult out-of-plane rotations. Moreover, the head pose yaw MAE on AFLW2000-3D is \textbf{2.92} for AC-DC vs. 3.29 for AC+Pose. Hence, using doubly-conditional fusion inside AC-DC architecture is beneficial for both landmark alignment and head pose estimation.

\begin{table}[ht]
    \small
    \centering
	\caption{Comparison in terms of NME (lower is better) on WFLW.}
	\label{ablation4}
	\begin{tabular}{l|r|r|r|r|r|r|r}
		\hline
		method&	all	&pose&expr&illu&m-up&occl&blur\\
		\hline
		\hline
		AC & 4.6 & 8.08 & 4.64 & 4.39 & 4.50 & 5.70 & 5.38 \\
		AC+Pose &4.55&7.86&4.61&\textbf{4.34}&4.31&5.66&5.33\\
		AC-DC 	&\textbf{4.49}&\textbf{7.76}&\textbf{4.45}&4.35&\textbf{4.25}&\textbf{5.57}&\textbf{5.21}\\
		\hline
	\end{tabular}
\end{table}
\subsection{2D face alignment}\label{2dal}

\begin{table}[ht]
    \small
	\caption{Comparison in terms of NME (lower is better), AUC (higher is better) as well as failure rate (lower is better), on WFLW.}
	\label{com1}
	\begin{tabular}{l|r|r|r|r|r|r|r}
		\hline
		method&all&pose&expr&illu&m-up&occl&blur\\
		\hline
		\multicolumn{8}{l}{NME (\%)}\\
		\hline
		LAB\cite{wu2018look}	&	5,27&	10,2&	5,51&	5,23&	5,15&	6,79&	6,32\\
		Wing\cite{feng2018wing} & 5.11 & 8.75 & 5.36 & 4.93 & 5.41 & 6.37 & 5.81\\
		3DDE\cite{valle2019face} & 4.68 & 8.62 & 5.21 & 4.65 & 4.60 & 5.77 & 5.41\\
		DeCaFA \cite{dapogny2019decafa}& 4.62 & 8.11 & 4.65 & 4.41 & 4.63 & 5.74 & 5.38\\

		\hline
		AC-DC&	\textbf{4.49}&	\textbf{7.76}&	\textbf{4.45}&	\textbf{4.35}&	\textbf{4.25}&	\textbf{5.57}&	\textbf{5.21}\\
		\hline
		\multicolumn{8}{l}{AUC@0.1 (\%)}\\
		\hline
		LAB\cite{wu2018look}	&	53.2&	23.5& 49.5&	54.3&	53.9&	44.9&	46.3\\
		Wing\cite{feng2018wing} & 55.4 & 31.0 & 49.6 & 541 & 55.8 & 48.9 & 49.2\\
		3DDE\cite{valle2019face} & 55.4 & 26.4 & 51.8 & 56.0 & 55.4 & \textbf{49.9} & 49.6\\
		DeCaFA \cite{dapogny2019decafa}& 56.3 & 29.2 & 54.6 & 57.9 & 57.5 & 48.5 & 49.4\\
		\hline
		AC-DC&	\textbf{57.5}&	\textbf{31.5}&	\textbf{56.6}&	\textbf{58.7}&	\textbf{58.3}&	49.5&	\textbf{51.1}\\
		\hline
		\multicolumn{8}{l}{FR@0.1 (\%)}\\
		\hline
		LAB\cite{wu2018look}	&	7.56&	28.8&	6.37&	6.73&	7.77&	13.7&	10,7\\
		Wing\cite{feng2018wing} & 6.00 & 22.7 & 4.78 & 4.30 & 7.77 & 12.5 & 7.76\\
		3DDE\cite{valle2019face} & 5.0 & 22.4 & 5.41 & 3.86 & 6.79 & 9.37 & 6.72\\
		DeCaFA \cite{dapogny2019decafa}& 4.84 & 21.4 & 3.73 & 3.22 & 6.15 & 9.26 & 6.61\\
		\hline
		AC-DC&	\textbf{4.29}&	\textbf{17.3}&	\textbf{2.69}&	\textbf{2.45}&	\textbf{4.66}&	\textbf{9.2}&	\textbf{5.82}\\
		\hline
	\end{tabular}
\end{table}

Next, we compare AC-DC with recent state-of-the-art approaches for 2D alignment on WFLW database. Most notably, AC-DC improves the landmark alignment accuracy on nearly every subset and metric, as compared to LAB \cite{wu2018look}, Wing \cite{feng2018wing} and 3DDE \cite{valle2019face}. Note that Wing \cite{feng2018wing} uses head pose to balance example sampling at train time and \cite{valle2019face} first infer head pose to pre-align a rigid set of landmarks before regressing the final extimates with coarse to fine ensemble of trees. By contrast, AC-DC jointly learns head pose and landmark alignment, each one benefiting from the other \textit{via} doubly conditional fusion, allowing to iteratively refine the predictions through the subsequent cascade stages. This, in turn, provides higher alignment accuracies. It should also be empathized that entwined head pose estimation, as well as the integration of the 3D landmark alignment task, heavily benefit the landmark localization on the \textit{pose} subset of WFLW in terms of NME, AUC as well as FR metric.

\subsection{3D face alignment}\label{3dal}

\begin{table}[ht]
    \centering
	\caption{Comparison in terms of Normalized Mean error (lower is better) on AFLW2000-3D.}
	\label{com2}
	\begin{tabular}{l|r|r|r|r}
		\hline
		method&	[0,30]	& [30,60] & [60,90] &avg\\
		\hline
		SDM\cite{Xiong2013} & 3.67 & 4.94 & 9.67 & 6.12\\
		3DSTN\cite{bhagavatula2017faster}&3.15&4.33&5.98&4.49\\
		3DDFA\cite{zhu2019face} 	&2.84&3.57&4.96&3.79\\
		PRN\cite{feng2018joint} 	&2.75&3.51&4.61&3.62\\
		2DASL\cite{tu2019joint} &2.75&3.44&\textbf{4.41}&3.53\\
		\hline
		AC-DC 	&\textbf{2.29}&\textbf{3.15}&4.76&\textbf{3.40}\\
		\hline
	\end{tabular}
\end{table}

\begin{figure*}[h!]
	\centering
	\includegraphics[width=1.0\linewidth]{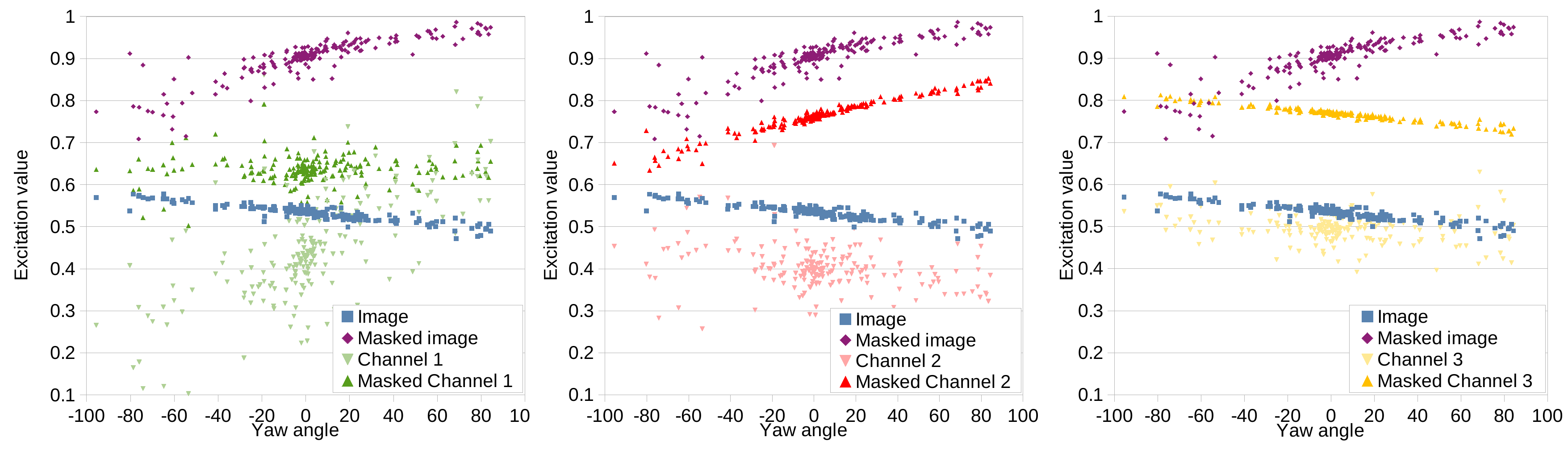}
	\caption{Excitation values for channels corresponding to the original image, masked image, embeddings and masked embeddings and the third cascade stage. from left to right: first, second and third embedding channels (as well as the corresponding masked embeddings).}
	\label{main5}
\end{figure*}

Next, in Table \ref{com2} we compare the accuracy of our method for 3D face alignment on AFLW2000-3D database. AC-DC significantly enhance the state-of-the-art results on the $[0,30]$ and $[30,60]$ head pose ranges. Furthermore, Despite the fact that both 3DDFA \cite{zhu2019face} and PRN \cite{feng2018joint} benefit from dense 3D morphable model fitting or UV map ground truth while AC-DC only aligns a sparse set of landmarks, our method achieves performances that are close to the state-of-the-art best approach, 2DASL \cite{tu2019joint} on the $[60,90]$ pose range. The average accuracy on those 3 pose bins is $\textbf{3.40}$ vs $3.53$ for 2DASL \cite{tu2019joint}. Note that 2DASL \cite{tu2019joint} also uses external 2D data for training. Also, the unweighted average accuracy (normalized error averaged on all images from AFLW2000-3D database without considering the pose subsets) is $\textbf{2.83}$ vs $3.07$ for the approach that use stacked dense U-Nets \cite{guo2018stacked}, which also integrates data from the MENPO dataset.

\subsection{Head pose estimation}\label{hppred}

\begin{table}[ht]
    \centering
	\caption{MAE on head pose estimation on AFLW-2000 3D. $^\dagger$: results excerpted from \cite{ruiz2018fine}.}
	\label{com3}
	\begin{tabular}{l|r|r|r|r}
		\hline
		method&	yaw	& pitch & roll &avg.\\
		\hline
		Trees$^\dagger$\cite{kazemi2014one}& 16.76 & 13.80 & 6.19 & 12.25\\
		FAN$^\dagger$\cite{bulat2017far}& 8.53 & 7.48 & 7.63 & 7.88\\
		3DDFA\cite{zhu2019face} &5.40&8.53&8.25&7.39\\
		Hopenet\cite{ruiz2018fine} &6.47&\textbf{6.56}&\textbf{5.44}&6.16\\
		\hline
		AC-DC (stage 1)	&3.48&8.26&6.87&6.20\\
		AC-DC (stage 4)	&\textbf{2.92}&6.94&5.99&\textbf{5.29}\\
		\hline
	\end{tabular}
\end{table}

Finally, in Table \ref{com3} we compare our approach with state-of-the-art methods for head pose estimation on AFLW2000-3D database. Our method performs better than Trees \cite{kazemi2014one} and FAN \cite{bulat2017far} which are a landmark-based methods. It is also better than 3DDFA \cite{zhu2019face} and the state-of the-art Hopenet \cite{ruiz2018fine}, which respectively uses 3D morphable model fitting and head pose estimation from the raw images without aligning facial landmarks. note that our method is most significantly better on the yaw angle estimation, which is the main benchmark on AFLW2000-3D. More precisely, after cascade stage 1, AC-DC is roughly as accurate as Hopenet \cite{ruiz2018fine} in terms of average head pose estimation whereas, after stage 4 it performs much better: this highlights the fact that using head pose information to refine landmark alignment provides more precise landmark estimates, which in turn helps refine the head pose prediction, further advocating for an entwined landmark alignment and head pose prediction scheme. 

\subsection{Qualitative analysis}\label{visus}

Figure \ref{main5} shows the excitation values (computed using a single dense layer from the estimated head pose, as illustrated on Figure \ref{main2}). For clarity purposes, we only plotted the excitation values for the first 200 examples of AFLW2000-3D database, and the first (left), second (middle) and third (right) channels of the embeddings (and of the corresponding masked embeddings). The conclusion of these charts are multiple: first, the excitation values for the original image and raw embeddings are significantly lower than the excitation values of their spatially-masked counterparts. This indicates that the network gives more weight to the spatially-masked channels to refine the landmark-wise attention maps in the subsequent cascade stages, showing the interest of designing a deep cascade as compared to a traditional deep approach. Second, the respective weights of the channels are heavily influenced by the yaw angle, notably for the masked image and channels. This influence is intrinsic to each individual channel, showing that channels are indeed selected using the head pose estimate and that the head pose conditionning takes place as expected.

\begin{figure*}[h]
	\centering
	\includegraphics[width=0.95\linewidth,height=0.59\linewidth]{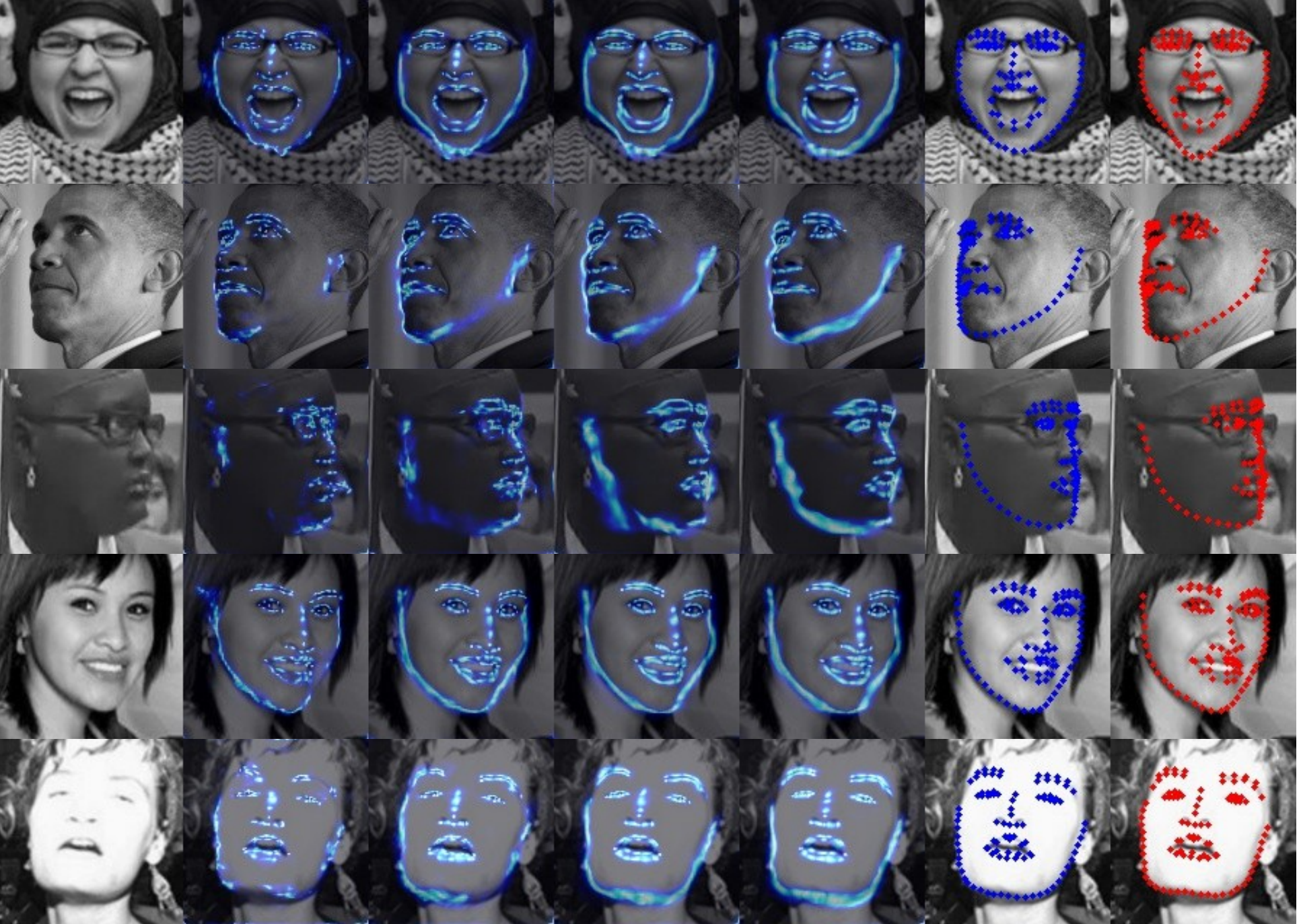}
	\caption{Examples of successful alignment on WFLW test set. From left to right: original image, fused attention maps after the first, second, third and fourth cascade stages, aligned and ground truth landmarks.}
	\label{main3}
\end{figure*}


Figures \ref{main3} shows examples of successful alignment on held out images from WFLW test set. Notice how the fused attention maps $\Phi_i(x,y)$ are coarse after the first and second cascade stages and are refined after the subsequent stages, outputting precise landmark alignment even under facial expression, non-planar head poses as well as difficult environmental lighting. 


Last but not least, note that landmark-wise attention maps are generally more spread-out in case of an occluded or misaligned landmark: as such, an interesting future direction would be to estimate landmark-wise alignment uncertainty from the spread measurement, which is possible under certain assumptions \cite{kendall2017uncertainties}.

\section{Conclusion}\label{concl}

In this paper, we proposed to entwine head pose estimation and facial landmark alignment tasks inside an attentional cascade. The proposed architecture employs a geometry transfer network (GTN), whose parameters are shared among the various cascade stages, to solve the annotation transfer task and integrate multiple heterogeneous landmark prediction tasks as well as head pose estimation within a single deep network. Our model also uses doubly-conditional fusion blocks to select relevant channels, and regions thereof, depending on a current head pose estimate, as well as attention maps corresponding to a current landmark alignment. The proposed AC-DC network can be trained in an end-to-end manner and significantly improves the state-of-the-art results for both 2D and 3D face alignment on recent large-scale datasets, as well as for head pose estimation, advocating for an entwined head pose estimation and face alignment pipeline. As such, AC-DC could be applied to closely related computer vision domains such as body pose estimation.

{\small
	\bibliographystyle{ieee}
	\bibliography{egbib}
}

\end{document}